\documentclass[journal,comsoc]{IEEEtran}
\usepackage[T1]{fontenc}% optional T1 font encoding
\usepackage{cite}

\ifCLASSINFOpdf
   \usepackage[pdftex]{graphicx}
\else
   \usepackage[dvips]{graphicx}
\fi

\usepackage{amsmath}
\usepackage[cmintegrals]{newtxmath}

\usepackage[numbers]{natbib}
\usepackage{booktabs}
\usepackage{multirow}
\usepackage{multicol}
\usepackage{amsmath}
\usepackage{algpseudocode}
\usepackage{mathtools}
\usepackage[linesnumbered,ruled,vlined]{algorithm2e}
\usepackage{url}

\usepackage{tikz}
\usetikzlibrary{arrows.meta, positioning, shapes.multipart, shapes.geometric}

\mathchardef\mhyphen="2D

\begin{document}
%
% paper title
% Titles are generally capitalized except for words such as a, an, and, as,
% at, but, by, for, in, nor, of, on, or, the, to and up, which are usually
% not capitalized unless they are the first or last word of the title.
% Linebreaks \\ can be used within to get better formatting as desired.
% Do not put math or special symbols in the title.
% \title{Scope expansion for LLM to increase the probability of the correctness of output}
\title{Human-Inspired Learning for Large Language Models via Obvious Record and Maximum-Entropy Method Discovery}

\author{Hong~Su
% <-this % stops a space
\IEEEcompsocitemizethanks{\IEEEcompsocthanksitem H. Su is with the School of Computer Science, Chengdu University of Information Technology, Chengdu, China.\\
 E-mail: suguest@126.com. \\
\protect\\
% note need leading \protect in front of \\ to get a newline within \thanks as
% \\ is fragile and will error, could use \hfil\break instead.
}% <-this % stops an unwanted space
\thanks{}}

% The paper headers
\markboth{Journal of \LaTeX\ Class Files,~Vol.~14, No.~8, August~2015}%
{Shell \MakeLowercase{\textit{et al.}}: Bare Demo of IEEEtran.cls for IEEE Communications Society Journals}
% The only time the second header will appear is for the odd numbered pages
% after the title page when using the twoside option.
%
% * Note that you probably will NOT want to include the author's *
% * name in the headers of peer review papers.                   *
% You can use \IFCLASSOPTIONpeerreview for conditional compilation here if
% you desire.

% make the title area
\maketitle

\begin{abstract}
Large Language Models (LLMs) excel at extracting common patterns from large-scale corpora, yet they struggle with rare, low-resource, or previously unseen scenarios—such as niche hardware deployment issues or irregular IoT device behaviors—because such cases are sparsely represented in training data. Moreover, LLMs rely primarily on implicit parametric memory, which limits their ability to explicitly acquire, recall, and refine methods, causing them to behave predominantly as intuition-driven predictors rather than deliberate, method-oriented learners.

Inspired by how humans learn from rare experiences, this paper proposes a \emph{human-inspired learning} framework that integrates two complementary mechanisms. The first, \emph{Obvious Record}, explicitly stores cause--result (or question--solution) relationships as symbolic memory, enabling persistent learning even from single or infrequent encounters. The second, \emph{Maximum-Entropy Method Discovery}, prioritizes and preserves methods with high semantic dissimilarity, allowing the system to capture diverse and underrepresented strategies that are typically overlooked by next-token prediction.

Verification on a benchmark of 60 semantically diverse question--solution pairs demonstrates that the proposed entropy-guided approach achieves stronger coverage of unseen questions and significantly greater internal diversity than a random baseline, confirming its effectiveness in discovering more generalizable and human-inspired methods.
\end{abstract}

% Note that keywords are not normally used for peerreview papers.

\IEEEpeerreviewmaketitle

\section{Introduction}

Large Language Models (LLMs) have achieved substantial progress across a wide 
range of reasoning, generation, and problem-solving tasks~\cite{chang2024survey}. 
Their training paradigm—predicting the next token over massive corpora—enables 
them to capture broad statistical regularities and to perform well on problems 
that are commonly represented in training data~\cite{ji2025overview}. 
Despite these strengths, LLMs exhibit significant limitations when confronted 
with \emph{rare, low-resource, or previously unseen scenarios}.

Typical examples include niche hardware deployment issues (e.g., uncommon GPU 
models), atypical IoT device failures, or real-world system problems that lack 
sufficient textual documentation online.  
For instance, mainstream software frameworks such as TensorFlow and PyTorch 
primarily provide default support for widely used GPU hardware and standard 
operating systems, whereas newly released or less common GPUs—especially when 
combined with highly customized or non-mainstream operating systems—often 
require device-specific configurations and undocumented adaptations.  
Because LLMs predominantly reflect commonly learned patterns, they are often 
ineffective when addressing such specialized GPU–OS combinations, forcing 
users to rely instead on targeted technical forums or vendor-specific 
documentation to obtain reliable solutions.  
As these cases are sparsely represented in training corpora, LLM-generated 
responses tend to be incomplete, inaccurate, or overly generic.

\begin{figure*}[!t]
    \centering
    \includegraphics[width=5in]{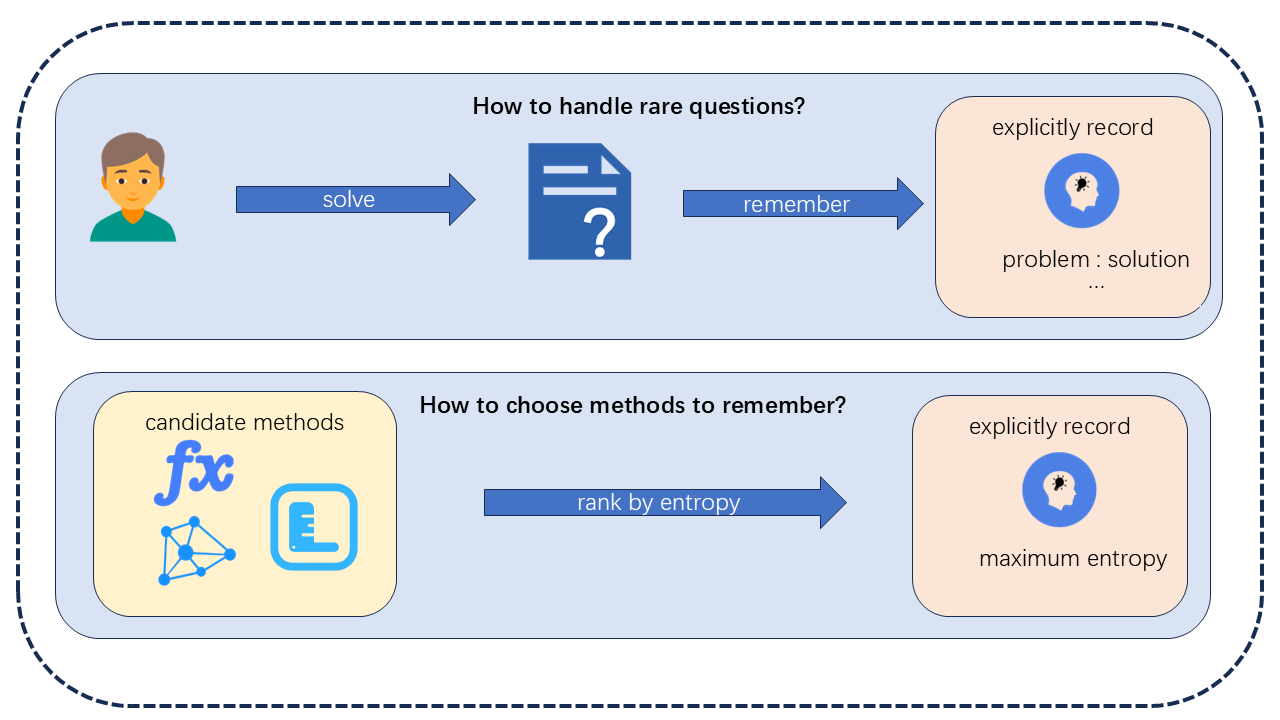}
    \caption{Overview of the proposed human-like learning framework. 
    Top: after solving a rare problem, the system explicitly stores a 
    \emph{problem$\rightarrow$solution} mapping in an Obvious Record for future reuse. 
    Bottom: when multiple candidate methods are available, MaxEn ranks them by 
    maximum semantic entropy and stores the most informative (high-entropy) methods, 
    thereby expanding semantic coverage while reducing redundancy.}
    \label{fig_overview}
\end{figure*}

A key reason for this limitation lies in the nature of parametric learning. 
In LLMs, knowledge is stored implicitly within weight matrices, and retrieval 
occurs through an intuition-like process in which the model selects 
high-probability continuations based on previously learned patterns. 
In contrast, human learners employ a dual mechanism: they rely on intuition 
for familiar situations while also maintaining \emph{explicit memory} of 
specific cause--result relationships, which enables them to recall rare 
experiences and refine methods over time. 
Such explicit memory is essential for handling infrequent events that 
intuition alone cannot resolve, as illustrated in Fig.~\ref{fig_overview}.

Motivated by this gap, we propose a human-like learning framework that augments
LLMs with two complementary capabilities:

\begin{itemize}
    \item \textbf{Obvious Record - Explicit Memory} --- an explicit, symbolic, non-parametric memory for storing mappings of the form
    \emph{feature$_{\text{cause}}$ → feature$_{\text{result}}$}.
    This mechanism enables the system to learn from single or rare encounters,
    preserve interpretable knowledge, and update methods when better solutions
    appear.

    \item \textbf{Maximum-Entropy Method Discovery} --- a mechanism for
    identifying and retaining methods that are \emph{most semantically
    different} from existing knowledge.
    These high-entropy methods capture diverse perspectives and novel
    strategies that LLMs tend to overlook because they are not reinforced by
    next-token prediction.
\end{itemize}

Together, these mechanisms form a dual-process learning model in which the LLM
acts as an intuition engine while the Obvious Record and entropy-guided
discovery form an explicit, continuously improving method memory.
This framework enables the system to better handle rare scenarios, reduce
over-reliance on common patterns, and learn in a manner closer to human
experience.

This paper makes the following contributions:
\begin{enumerate}
    \item We propose a \textbf{Maximum-Entropy Method Discovery} strategy that 
    identifies semantically diverse and underrepresented methods, prioritizing 
    those most valuable for addressing novel or rare problems.

    \item We introduce a novel \textbf{Obvious Record with Entropy} mechanism 
    that provides explicit and interpretable memory for storing and refining 
    cause--result relationships, including results with either higher 
    effectiveness or higher semantic entropy, thereby enabling one-shot and 
    few-shot learning without modifying model parameters.

    \item We develop a \textbf{group-based entropy measurement} that quantifies 
    the semantic difference between a new candidate method and a set of 
    previously learned methods, enabling principled selection and integration 
    of genuinely novel knowledge.
\end{enumerate}

The remainder of this paper is organized as follows.
Section~\ref{sec_rel_wor} reviews related work on LLM learning mechanisms, memory augmentation, and diversity-based selection strategies.
Section~\ref{sec_model} presents the proposed human-like learning model, including the
Obvious Record mechanism, Maximum-Entropy Method Discovery, and the complete
learning pipeline.
Section~\ref{sec_verification} reports the verification experiments and provides a detailed analysis of the results.
Section~\ref{sec_con} concludes the paper and outlines directions for future research.

\section{Related Work} \label{sec_rel_wor}

This section reviews prior studies relevant to the proposed human-like learning
framework, including (i) limitations of parametric learning in LLMs,
(ii) explicit and external memory mechanisms,
(iii) diversity- and entropy-based learning strategies, and
(iv) method-learning and reasoning frameworks.
We highlight where existing approaches fall short and how our model achieves a
more human-like ability to learn from rare or unseen scenarios.

\subsection{Parametric Learning and Its Limitations}

LLMs such as GPT, PaLM, and LLaMA learn primarily through large-scale
next-token prediction~\cite{brown2020language, chowdhery2022palm}.
This training paradigm captures high-frequency patterns and enables impressive
zero-shot generalization, but it encodes knowledge implicitly inside model
weights.
As a result, LLMs tend to struggle with tasks that require retrieving
\emph{rare}, \emph{specific}, or \emph{non-distributed} experiences, such as
uncommon system configurations, novel IoT faults, or niche software issues.
Recent works have highlighted that LLMs often fail on low-resource domains due
to the absence of explicit, symbolic memory mechanisms~\cite{min2023fact}.

Several studies reveal that LLM predictions resemble ``intuition'' rather than
explicit method recall~\cite{su2025intuition}.
This limits their ability to refine methods over time or make use of previous
failures---behaviors that human learners naturally exhibit.
The proposed Obvious Record mechanism directly addresses this limitation by
storing interpretable cause--result mappings outside the model parameters.

\subsection{External Memory and Retrieval-Augmented Models}

To mitigate the limitations of purely parametric learning, retrieval-augmented
models (RAG) incorporate external documents during inference~\cite{lewis2020rag}.
Other works introduce memory modules~\cite{graves2014ntm, sukhbaatar2015memnn}
that allow the model to read and write external information.

However, these approaches primarily provide access to factual or textual
knowledge rather than structured \emph{methods}.
Furthermore, they lack mechanisms for determining which new information should
be stored or how to refine stored knowledge over time.
Our framework differs in that:

\begin{itemize}
    \item it records \emph{methods} (cause $\rightarrow$ result), not facts;
    \item it uses entropy-based novelty detection to decide what to store;
    \item it supports continuous improvement of stored methods.
\end{itemize}

Thus, the Obvious Record serves as a method-specific external memory,
complementing rather than replacing existing retrieval systems.

\subsection{Diversity and Entropy in Learning Systems}

Diversity-based sampling and entropy-driven learning have been widely studied
in clustering, active learning~\cite{ren2021survey},
and contrastive representation learning~\cite{chen2020simclr}.
These methods generally aim to improve coverage of the underlying data
distribution by selecting points that are dissimilar to previous selections.

In natural language processing, semantic diversity has been used for data
augmentation and example selection~\cite{yuan2021few}, but existing approaches
do not support explicit symbolic storage of high-entropy methods or
integration with human-like reasoning pipelines.

Our Maximum-Entropy Method Discovery differs in three key ways:

\begin{itemize}
    \item it operates in the space of \emph{methods}, not data points;
    \item it decides when to preserve a new method based on semantic novelty;
    \item it extracts top-$k$ distinctive features (EnEx-$k$) for compact,
    interpretable storage.
\end{itemize}

These innovations enable the acquisition of rare, high-impact methods that
LLMs typically overlook.

\subsection{Method Learning and Symbolic Reasoning}

Prior work has explored structured method learning, including the
question--solution method representation in~\cite{su2025method},
program induction~\cite{ellis2021dreamcoder}, and symbolic reasoning with
neural models~\cite{dong2019neural}.
Such works show the importance of explicit representations for interpretability
and compositional reasoning.

However, existing models generally assume:

\begin{itemize}
    \item abundant training examples for each method, or
    \item fixed symbolic reasoning rules,
\end{itemize}

neither of which holds in real-world problem-solving where many methods appear
only once.
Our framework extends prior method-learning work by introducing
entropy-guided novelty detection and continuous improvement mechanisms,
allowing the system to evolve its method repertoire dynamically based on
experience.

\subsection{Summary}

Existing LLMs lack explicit method memory and struggle with rare scenarios;
retrieval systems store facts but not structured methods; diversity-based
approaches do not perform symbolic method storage; and prior method-learning
frameworks require richer supervision.
The present work integrates these threads into a unified, human-like learning
model capable of:

\begin{itemize}
    \item storing methods explicitly,
    \item discovering diverse high-entropy strategies,
    \item refining methods through continuous comparison, and
    \item adapting to rare or previously unseen tasks.
\end{itemize}

\section{The Human-Inspired Learning Model} \label{sec_model}

\subsection{Overview of the Human-Inspired Learning Model}

Large Language Models (LLMs) primarily rely on parametric learning, where
knowledge is encoded implicitly inside weight matrices through next-token
prediction. This enables strong intuition-like reasoning but limits the
model’s ability to acquire, refine, or preserve explicit methods—especially
for rare or previously unseen scenarios. In contrast, human learning often
combines two complementary processes: (i) intuitive pattern recognition and
(ii) explicit recording of methods and outcomes that can be recalled and
improved over time.

Inspired by this dual-process behavior, we propose a human-like learning model
that augments LLMs with two mechanisms:

\begin{enumerate}
    \item \textbf{Obvious Record}: an explicit non-parametric memory that
    stores symbolic mappings from causes to results
    (e.g., question $\rightarrow$ solution, scenario $\rightarrow$ action).
    This allows the system to retain knowledge even from single encounters and
    to refine stored methods when better solutions are discovered.

    \item \textbf{Maximum-Entropy Method Discovery}: a mechanism for
    identifying and preserving methods that are semantically
    \emph{most dissimilar} from existing knowledge. These high-entropy methods
    represent diverse strategies that are typically underrepresented in LLM
    training data and are essential for solving rare or novel problems.
\end{enumerate}

Figuratively, the proposed framework positions the LLM as the
``intuition engine'' while the Obvious Record and entropy-based method
discovery operate as an explicit ``method memory'' that supports continuous
improvement. When a new task arrives, the system extracts key features from
the input, measures their semantic entropy against the existing record set,
and decides whether the new information should be recorded as a distinct
method. During reasoning, the system retrieves either the closest matching
method or, when existing methods fail, the highest-entropy alternative to
promote diverse problem-solving behavior.

This integrated design enables human-like adaptability: the model can learn
from infrequent events, preserve diverse strategies, refine outdated methods,
and avoid over-reliance on common patterns learned from large corpora.

\subsection{Obvious Record: Explicit Cause--Result Memory}

LLMs typically encode knowledge implicitly within high-dimensional parameter
spaces, which makes it difficult to explicitly recall, refine, or update
individual methods.
In contrast, human learners often retain task-specific experiences in an
explicit form, such as remembering that ``when situation $A$ occurs, action
$B$ works best.''
To emulate this capability, we introduce the \emph{Obvious Record}, a symbolic
and non-parametric memory that stores knowledge in the form of structured
cause--result mappings.

Formally, an Obvious Record is defined as:
\begin{equation}
    record:\; feature_{\text{cause}} \rightarrow feature_{\text{result}},
    \label{eq:obvious_record_mapping}
\end{equation}
where the cause feature represents a situation or question, and the result
feature represents the corresponding action or solution.
This formulation generalizes previously proposed method-learning schemes that
store question--solution pairs~\cite{su2025method}, and it naturally extends to
scenario--action relationships commonly encountered in practical environments,
such as IoT systems.

Notably, $feature_{\text{result}}$ is not restricted to a single element and
may instead be a set, since multiple valid methods can exist for addressing
the same cause; for example, a specific question may admit several alternative
solution strategies.

All Obvious Records are stored in a dedicated memory structure, referred to as
the \emph{human-like learning set}, which remains persistent, interpretable,
and independent of the LLM’s parametric weights.

\subsubsection{Structure of the Obvious Record}
Obvious Records can naturally form hierarchical or relational structures.
When multiple issues share a common $feature_{\text{cause}}$ but differ in
additional attributes, the records can be organized as a tree, where the root
node represents the common cause and child nodes encode more specific
conditions.
For instance, ``fire suppression'' may serve as a parent cause, while
``electrical fire'' forms a child node that introduces additional constraints
on feasible actions.

More generally, when a $feature_{\text{cause}}$ partially overlaps with or
subsumes another cause feature, the resulting relationships form a graph
structure rather than a strict hierarchy.
This flexible organization allows Obvious Records to represent complex
dependencies and conditional reasoning patterns encountered in practical
problem-solving.

\subsubsection{Recording Procedure}

When a new scenario or question is encountered, the system first extracts the
salient semantic features from the input (see Section~\ref{sec:entropy} for
details).
If the extracted $feature_{\text{cause}}$ has not been previously observed in
the human-like learning set, the system creates a new Obvious Record according
to~\eqref{eq:obvious_record_mapping}.

This explicit storage mechanism is particularly important in settings where
fine-tuning or retraining an LLM is impractical, such as IoT edge devices that
operate under limited data availability, restricted computational resources,
or rare and irregular event patterns.
Under these conditions, the Obvious Record enables effective one-shot or
few-shot learning without modifying the underlying model parameters.

\subsubsection{Continuous Improvement Mechanism}

Human learners do not merely store experiences; they continuously refine them
based on outcomes.
To emulate this behavior, the Obvious Record incorporates a continuous
improvement mechanism: when multiple results are associated with the same or
similar $feature_{\text{cause}}$, the system evaluates their effectiveness and
updates the stored record accordingly.

Let $r_a$ and $r_b$ denote two candidate results corresponding to the same cause
feature $c$.
The memory update rule is defined as:
\begin{equation}
    record(c) =
    \begin{cases}
        r_a, & \text{if } \text{eval}(r_a) > \text{eval}(r_b), \\[4pt]
        r_b, & \text{if } \text{eval}(r_b) > \text{eval}(r_a), \\[4pt]
        \{r_a, r_b\}, & \text{if } \text{eval}(r_a) = \text{eval}(r_b),
    \end{cases}
    \label{eq:continuous_improvement}
\end{equation}
where $\text{eval}(\cdot)$ denotes a task-dependent evaluation function, such as
correctness, utility, robustness, or execution stability.

This mechanism allows the system to progressively upgrade its stored methods
whenever superior solutions are discovered, closely mirroring how humans
refine problem-solving strategies over time.
When newly observed information contradicts an existing record, the system
either replaces or augments the outdated result, ensuring that the Obvious
Record consistently reflects the most effective known methods.

\subsection{Maximum-Entropy Method Discovery}
\label{sec:entropy}

Obvious learning provides a mechanism for recording learned methods, but an
equally important issue lies in deciding \emph{what} content should be learned,
because in practical environments many events occur repeatedly and convey
little new information, whereas only a small portion corresponds to genuinely
novel knowledge that has not been encountered before.
For example, when a new issue arises, the system may need to learn a new
$feature_{\text{cause}}$ (such as a previously unseen question) or a new
$feature_{\text{result}}$ (such as a solution method that has not been applied
before), rather than repeatedly recording similar questions with similar
solutions.
In such situations, it is neither necessary nor efficient to record all experiences indiscriminately.
Motivated by this observation, we propose prioritizing the learning of
information that is \emph{new} relative to existing knowledge, where novelty is
quantified using semantic entropy.

While the Obvious Record mechanism enables explicit storage of experiences, it does not by itself determine which experiences are sufficiently important or distinctive to be preserved, whereas humans naturally retain events that are highly informative, particularly when previously learned knowledge fails to provide an effective solution.
To emulate this selective aspect of human learning, we introduce
\emph{Maximum-Entropy Method Discovery}, a mechanism designed to identify and preserve methods that are semantically most different from those already stored in the human-like learning set.

In this context, entropy is used to measure semantic \emph{dissimilarity}, which we operationalize through cosine distance computed either from an LLM-based embedding model or from conventional vector representations, and in this work we adopt an LLM embedding model because its semantic space aligns more closely with human judgments of meaning.
Given two semantic vectors $A$ and $B$, the cosine-distance entropy is defined as:
\begin{equation}
    EN_{\text{cos}}(A, B) = 1 - S_{\text{cos}}(A, B),
    \label{eq:entropy_define}
\end{equation}
where $S_{\text{cos}}$ denotes their cosine similarity.
A higher value of $EN_{\text{cos}}$ indicates that $A$ and $B$ encode
substantially different meanings, and methods associated with high entropy are therefore especially valuable because they represent alternative perspectives or novel strategies that are typically overlooked by the LLM’s intuition-driven next-token prediction process.

\subsubsection{Group Entropy}

To determine whether a newly encountered method is sufficiently novel, it is
necessary to compare it not with a single existing record but with the entire
human-like learning set, since learning in practice is always based on a
collection of previously acquired methods rather than on isolated examples.
Accordingly, we introduce the notion of \emph{group entropy}, which measures
both the diversity within an existing set of learned methods and the novelty
of a new method relative to that set.

(a) Internal Group Entropy.
For a set $S = \{S_1, S_2, \ldots, S_m\}$, we define its internal entropy as:
\begin{equation}
    EN_{\text{internal}}(S) =
    \max_{i < j} EN_{\text{cos}}(S_i, S_j),
    \label{eq:internal_entropy}
\end{equation}
which measures the maximum semantic distance between any two members of the
set.
A high internal entropy indicates that the learned methods in $S$ are widely
diverse and cover substantially different semantic regions.

(b) External Entropy of a New Item.
When a new feature $A$ is compared against the existing set $S$, we compute:
\begin{equation}
    EN_{\text{external}}(A, S) =
    \min_{i} EN_{\text{cos}}(A, S_i),
    \label{eq:external_entropy}
\end{equation}
which represents the semantic distance between $A$ and its closest existing
record.
A high value of $EN_{\text{external}}(A, S)$ implies that $A$ introduces novel
information that is not covered by the current knowledge stored in the human-like learning set.

\subsubsection{High-Entropy Learning Rule}

To emulate human selective memory, a newly encountered method is recorded only
when its semantic entropy exceeds a predefined threshold $\tau$, ensuring that
the learning process prioritizes genuinely novel information rather than
redundant variations.

Specifically, a new cause--result mapping is added to the Obvious Record if:
\begin{equation}
    EN_{\text{external}}(feature_{\text{cause}}, S) \ge \tau,
    \label{eq:high_entropy_threshold}
\end{equation}
which indicates that the input corresponds to a substantially different
scenario or problem compared with existing knowledge.
Similarly, for solution methods, a new result is considered novel if:
\begin{equation}
    EN_{\text{external}}(feature_{\text{result}}, S) \ge \tau,
    \label{eq:high_entropy_threshold_result}
\end{equation}

This criterion ensures that only high-entropy knowledge expands the memory,
while low-entropy items—being semantically close to existing records—are
treated as variations of known cases and do not enlarge the record set.

Importantly, for the same cause feature $c$, multiple solution methods may be
retained simultaneously when they are semantically different.
That is, if two candidate results $r_a$ and $r_b$ both satisfy the entropy
criterion and exhibit high semantic dissimilarity, they are jointly preserved
even if one achieves a higher evaluation score than the other.
This reflects the human tendency to remember multiple distinct strategies for
the same problem rather than collapsing them into a single ``best'' solution.

Accordingly, the continuous improvement rule is refined as follows:
\begin{equation}
    record(c) =
    \begin{cases}
        \{r_a, r_b\}, & \text{if } EN_{\text{cos}}(r_a, r_b) \ge \tau, \\[4pt]
        r_a, & \text{if } EN_{\text{cos}}(r_a, r_b) < \tau
        \ \text{and } \text{eval}(r_a) > \text{eval}(r_b), \\[4pt]
        r_b, & \text{if } EN_{\text{cos}}(r_a, r_b) < \tau
        \ \text{and } \text{eval}(r_b) > \text{eval}(r_a).
    \end{cases}
    \label{eq:continuous_improvement}
\end{equation}

This rule balances effectiveness and diversity: semantically distinct methods
are preserved to support exploration, while similar methods are refined based
on task-dependent performance.

\subsubsection{Top-$k$ Entropy Extraction (EnEx-$k$)}
Directly comparing long textual inputs may dilute semantically critical
differences.
To focus on the distinguishing elements, we introduce the
\emph{Top-$k$ Entropy Extraction} (EnEx-$k$) mechanism.

Given an input text, we extract the $k$ features (typically words or phrases)
that contribute most strongly to its entropy relative to existing records.
These features serve as compact representations of
$feature_{\text{cause}}$ or $feature_{\text{result}}$.

This yields a concise record representation:

\begin{equation}
    record:\; \text{top-}k(feature_{\text{cause}})
    \rightarrow
    \text{top-}k(feature_{\text{result}}),
    \label{eq:record_topk}
\end{equation}

analogous to how humans remember only the key distinctive aspects of an event
rather than all details.
In most applications, $k=1$ or $k=2$ captures the essential semantic
difference while avoiding unnecessary noise.

\subsection{Method Selection with Maximum Entropy}

Once the Obvious Record and the entropy-based discovery mechanism are
established, the system must determine how to select an appropriate method
when solving a new problem.
Human problem-solving provides a useful analogy: people typically rely on
familiar methods first, but when these methods fail, they deliberately seek
alternative perspectives that differ significantly from prior experience.
Our model formalizes this behavior through entropy-guided method selection.

Given an input query with extracted cause features $c_{\text{new}}$, the system
operates under two complementary reasoning modes when a question is difficult
to solve, for example when a user or the system has attempted multiple times
without success:

\begin{enumerate}
    \item \textbf{Similarity-Based Retrieval (Routine Mode):}
    When the new problem closely resembles previously encountered cases, the
    system retrieves the method whose cause feature is most similar to
    $c_{\text{new}}$.
    This mode reflects intuition-driven or habitual reasoning, in which known
    methods are reused efficiently.
    However, if the retrieved method fails to produce a correct solution for
    the target $c_{\text{new}}$, the failure may indicate that the underlying
    cause feature has not been correctly identified.
    In such cases, if an alternative $c_{\text{new}}$ with sufficiently high
    entropy is detected, the system treats it as a potentially new issue and
    updates the cause representation accordingly.

    \item \textbf{Entropy-Based Retrieval (Exploration Mode):}
    When previously applied methods in $feature_{\text{result}}$ fail to solve
    the problem, or when the system is explicitly instructed to explore
    alternatives, it selects a new method that is maximally different from
    those already attempted.
    Let $methods_{\text{tried}}$ denote the set of unsuccessfully applied
    methods; the next method is chosen as:
    \begin{equation}
        r^{*} =
        \arg\max_{r \in S}
        EN_{\text{cos}}(methods_{\text{tried}}, r),
        \label{eq:select_max_entropy}
    \end{equation}
    which ensures that the next attempted solution differs as much as possible
    from all previously tried methods.
\end{enumerate}

This dual-mode selection strategy mirrors human reasoning behavior: routine
methods are applied when appropriate, while high-entropy methods introduce
substantially different solution paths when familiar approaches prove
insufficient.

\subsubsection{Why High-Entropy Methods Are Valuable}

If a low-entropy method fails, it typically indicates that the new problem
differs in essential ways from previously encountered situations.
Therefore, selecting another low-entropy method—one similar to the failed
approach—is unlikely to succeed.

Let $m_{\text{fail}}$ be the failed method and $m_i$ be another candidate
method.
If:
\begin{equation}
    EN_{\text{cos}}(m_i, m_{\text{fail}}) \approx 0,
\end{equation}
then $m_i$ is semantically close to the failed method and is unlikely to
produce a substantially different outcome.

Conversely, a high-entropy method satisfies:
\begin{equation}
    EN_{\text{cos}}(m_i, m_{\text{fail}}) \gg 0,
\end{equation}
indicating that it represents a significantly different strategy.
Thus, high-entropy methods offer new solution directions, analogous to how
humans seek different viewpoints or alternative heuristics when stuck.
\subsubsection{Retrieval and Ranking of Methods}

The complete retrieval process for a query with extracted cause feature
$c_{\text{new}}$ proceeds as follows:

\begin{enumerate}
    \item Compute the semantic entropy or similarity between $c_{\text{new}}$
    and each stored $feature_{\text{cause}}$ in the human-like learning set $S$.
    \item Rank the candidate methods according to the active reasoning mode,
    using either semantic similarity or semantic entropy as the ranking
    criterion.
    \item Select the most appropriate candidate:
    \begin{itemize}
        \item the method with the highest similarity in routine (similarity-based) reasoning, or
        \item the method with the highest entropy in exploratory (entropy-based) reasoning.
    \end{itemize}
    \item Apply the selected method; if the outcome is ineffective, the system
    transitions to entropy-based retrieval and considers high-entropy
    alternatives.
\end{enumerate}

This ranking-based retrieval strategy yields an interpretable and structured
decision-making process, helping to mitigate the common ``black-box'' criticism
associated with LLM-based systems.

\subsubsection{Identifying Distinct Sub-Problems}

A single user query may implicitly involve multiple underlying sub-problems,
some of which may be conceptually independent.
Entropy-based analysis provides a principled way to detect such distinctions
by measuring semantic dissimilarity between extracted cause features.

Given two extracted cause features $c_1$ and $c_2$, if:
\begin{equation}
    EN_{\text{cos}}(c_1, c_2) \ge \tau,
\end{equation}
the system treats them as independent sub-problems and retrieves or records
methods for each separately.

% This capability is particularly valuable for complex diagnostic scenarios, such as software debugging or IoT system maintenance, where multiple unrelated issues may arise simultaneously and require distinct solution strategies.

\section{Verification} \label{sec_verification}

This section evaluates whether Maximum-Entropy Method Discovery (MaxEn)
provides superior semantic coverage and method diversity compared with a
random-choice baseline (RanCho).
Since MaxEn is intended to help the system learn diverse and human-like
methods, we measure two key properties:

\begin{enumerate}
    \item \textbf{External Coverage:}
    Measures how close the selected methods are to a randomly sampled, unseen question. This evaluates the effectiveness of the proposed learning mechanism in covering the broader semantic space.

    \item \textbf{Internal Diversity:}
    Measures how semantically different the selected methods are from one another. This reflects the internal structure of the learned method set and indicates whether the model captures a diverse range of strategies.
\end{enumerate}

A curated benchmark of semantically diverse question--solution pairs is used to
simulate the kinds of conceptual methods that a human or LLM-based learner
might accumulate over time.

\subsection{Verification Setup}

We employ a benchmark dataset consisting of 60 question--solution pairs,
denoted QSS60, which is publicly available on Zenodo~\cite{qss60}.
Each pair represents a distinct semantic region spanning a variety of domains
including software engineering, machine learning, blockchain, IoT systems,
experimental design, and high-level reasoning.
Embeddings are generated using the
\texttt{distiluse-base-multilingual-cased-v1} model to simulate the semantic
understanding used in human-like learning.

The goal is to evaluate whether MaxEn is more effective than RanCho at
constructing representative subsets of size:

\[
n \in \{2,4,6,8,10,12,14\},
\]

corresponding to different levels of accumulated experience.

\subsection{Compared Strategies}

Two strategies are compared when selecting $n$ items from QSS60:

\begin{itemize}
    \item \textbf{MaxEn (Entropy-Maximizing Selection).}
    Builds the subset greedily by repeatedly selecting the question with the
    \emph{least similarity} (i.e., highest semantic entropy) to all previously
    chosen ones.
    This approximates maximizing the diversity of learned methods.

    \item \textbf{RanCho (Random Choice Baseline).}
    Uniformly samples $n$ questions without replacement.
    This serves as a naive baseline lacking semantic reasoning.
\end{itemize}

% \subsubsection{Semantic Similarity Model}

For any pair of questions $q_i$ and $q_j$, semantic similarity is computed as:

\begin{equation}
    \mathrm{sim}(q_i, q_j) =
    \frac{\langle e_i, e_j \rangle}
         {\|e_i\| \cdot \|e_j\|},
\end{equation}

where $e_i$ and $e_j$ are their normalized embeddings.
The value ranges from $0$ (unrelated) to $1$ (nearly identical).
Entropy is implicitly captured by cosine distance:
$\mathrm{distance} = 1 - \mathrm{sim}$.

% \subsubsection{Evaluation Protocol}

Two evaluation tracks are employed:

\paragraph{Track 1 - External Similarity Test}

A question is sampled uniformly from QSS60 and the procedure is repeated 20 times.
For each selected subset, we compute the maximum semantic similarity between
the sampled question and the selected items in order to evaluate whether the
learned methods can effectively cover potential future issues.
A higher similarity value indicates stronger semantic coverage, meaning that
the selected methods better span the underlying conceptual space.

\paragraph{Track 2 - Internal Similarity Sum}

For each selected subset, we compute:

\begin{equation}
S = \sum_{i < j} \mathrm{sim}(q_i, q_j),
\end{equation}

which measures the degree of internal clustering. A lower $S$ corresponds to greater semantic diversity, meaning the selected questions cover a wider range of conceptual methods.

\subsection{Verification Results} \label{sec:verification_results}

The external similarity test is repeated 20 times with independent draws from QSS60.  Reported values are arithmetic means across trials. Internal similarity is deterministic once a subset is selected, resulting in one value per strategy for each $n$.

The results are summarized in this subsection.
Overall, MaxEn consistently achieves both higher external coverage and lower
internal similarity than RanCho across all subset sizes, demonstrating that
entropy-guided selection leads to more diverse and representative method
collections.

\subsubsection{External Similarity to a Randomly Selected Question}

\begin{figure}[!t]
    \centering
    \includegraphics[width=3.5in]{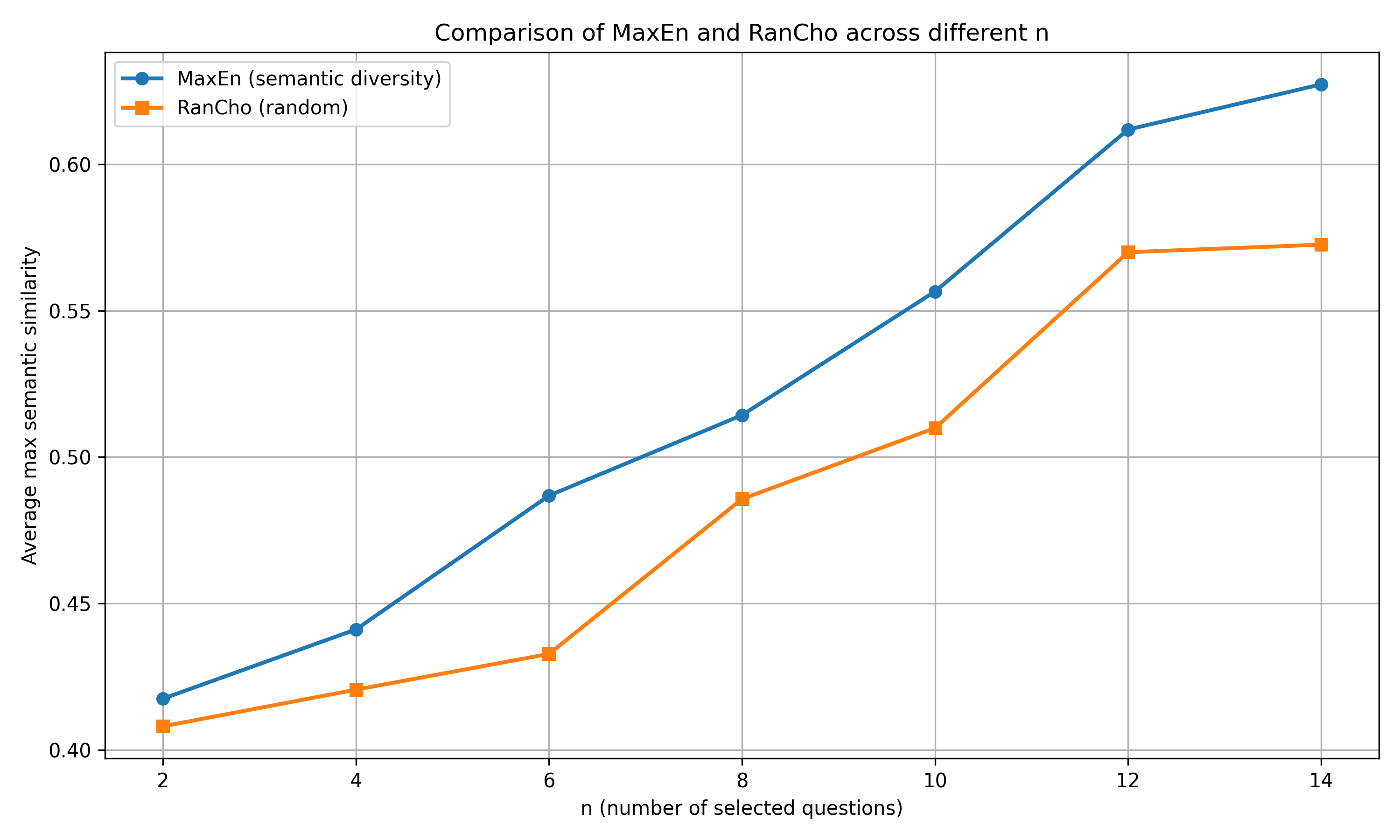}
    \caption{External similarity comparison between MaxEn and RanCho.}
    \label{comparison_plot}
\end{figure}

Figure~\ref{comparison_plot} and Table~\ref{tab:external_similarity} show that
MaxEn consistently achieves higher maximum similarity to randomly sampled,
unseen questions than RanCho.
For example, at $n=10$, MaxEn attains a similarity of 0.5566 compared with
RanCho’s 0.5099, a difference of 0.0467.
This trend holds across all tested values of $n$, indicating that
entropy-guided selection provides stronger \emph{semantic coverage} of the
overall question space.

\begin{table}[!t]
\centering
\caption{Average maximum similarity to a randomly selected question (20 trials).
$\Delta = \text{Avg\_MaxEn} - \text{Avg\_RanCho}$.}
\label{tab:external_similarity}
\begin{tabular}{c|c|c|c}
\hline
\textbf{$n$} & \textbf{Avg\_MaxEn} & \textbf{Avg\_RanCho} & \textbf{$\Delta$} \\
\hline
2  & 0.4175 & 0.4081 & 0.0094 \\
4  & 0.4412 & 0.4206 & 0.0206 \\
6  & 0.4869 & 0.4328 & 0.0541 \\
8  & 0.5143 & 0.4857 & 0.0286 \\
10 & 0.5566 & 0.5099 & 0.0467 \\
12 & 0.6118 & 0.5700 & 0.0418 \\
14 & 0.6273 & 0.5725 & 0.0548 \\
\hline
\end{tabular}
\end{table}

Although MaxEn prioritizes internal diversity during subset construction, the
resulting sets are distributed in such a way that an unseen question is more
likely to be semantically close to at least one selected item.
This demonstrates that learning with maximum entropy improves the model’s
generalization ability when both strategies acquire the same number of
methods.
To quantify this advantage, Table~\ref{tab:external_similarity} reports the
performance gap $\Delta = \text{Avg\_MaxEn} - \text{Avg\_RanCho}$ for each
subset size, with a max value of 0.0548.

The performance gap generally increases with $n$, suggesting that MaxEn
benefits from larger subset sizes, where diversity constraints have greater
impact.
In contrast, RanCho provides no guarantee of semantic spacing, leading to
weaker alignment with unseen questions as the subset grows.

\subsubsection{Internal Similarity Sum}

\begin{figure}[!t]
    \centering
    \includegraphics[width=3.5in]{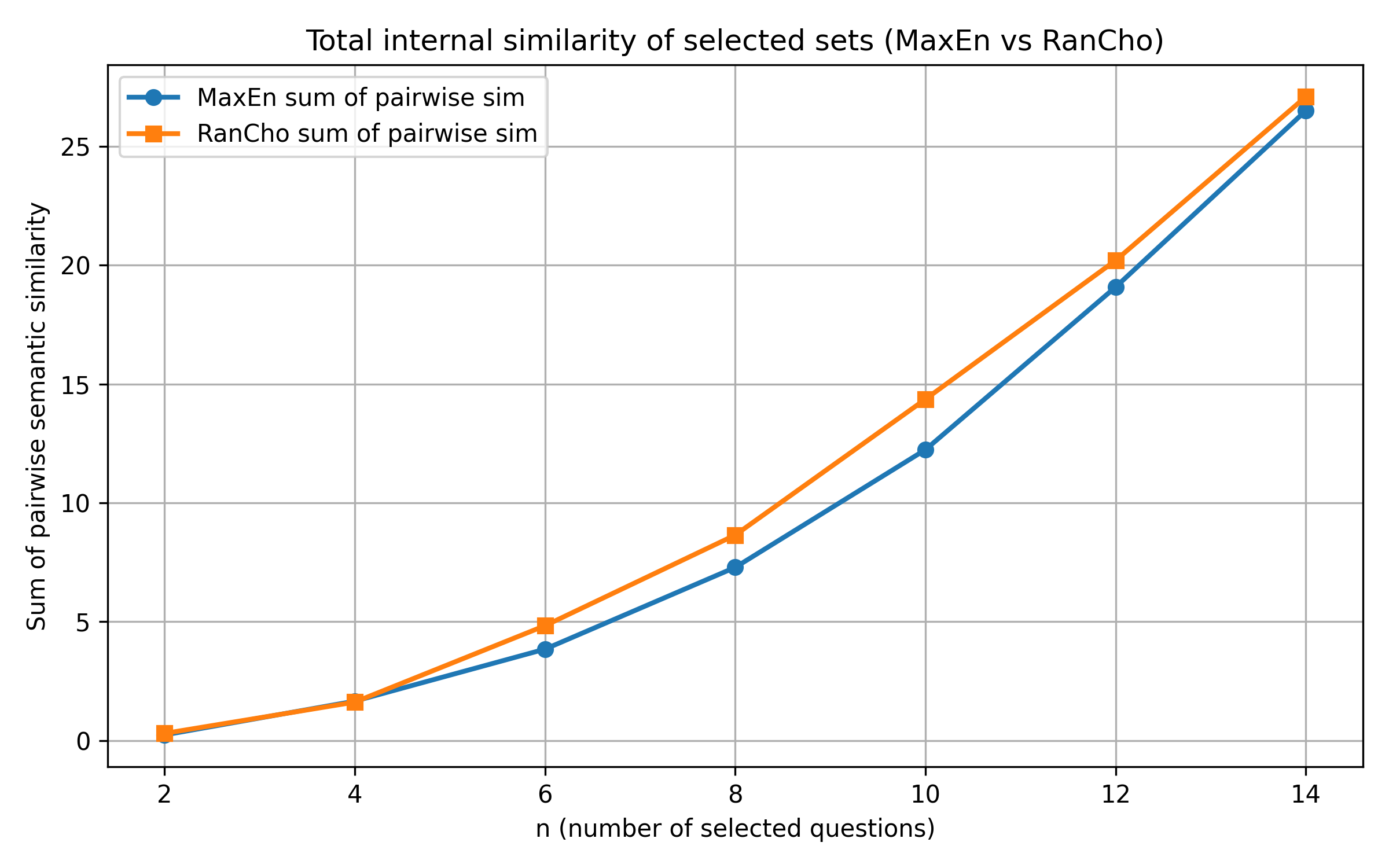}
    \caption{Sum of pairwise semantic similarities within selected subsets.}
    \label{sum_pairwise_similarity}
\end{figure}

Figure~\ref{sum_pairwise_similarity} shows that MaxEn consistently produces an internal similarity sum that is never higher than that of RanCho across all tested values of $n$. Most MaxEn values lie noticeably below the corresponding RanCho values, particularly when $n > 4$. For example, when $n = 10$, MaxEn yields an internal similarity sum of 12.23, compared with RanCho’s 14.36 - a difference of more than 2.

These results indicate that entropy-guided learning yields method sets that are
more diverse and more evenly distributed within the embedding space.
In contrast, RanCho exhibits higher and more variable similarity sums because
uniform random sampling provides no structural guarantee of semantic spacing.
% The reason is that MaxEn explicitly selects each new item to be least similar to all
% previously chosen ones, the resulting subsets maintain strong semantic
% dispersion and effectively avoid redundancy.

\par\vspace{0.5\baselineskip}

Taken together with the external similarity results, the evidence shows that
MaxEn supports more human-like method acquisition by achieving strong semantic
coverage while preserving conceptual variety.
This enables the system to adapt more effectively to new and unfamiliar
scenarios with fewer redundant methods.

\section{Conclusion} \label{sec_con}

This paper introduced a human-inspired learning framework that augments Large
Language Models with explicit symbolic memory and entropy-guided method
discovery. The proposed \emph{Obvious Record} mechanism provides a structured
non-parametric memory for storing and refining cause - result relationships,
allowing the system to learn effectively even from single or rare encounters.
Complementing this, the \emph{Maximum-Entropy Method Discovery} mechanism
identifies and preserves semantically diverse methods that are often
underrepresented in traditional LLM training. Together, these components form
a dual-process learning architecture that more closely mirrors human
experience: routine reasoning is supported by similarity-based retrieval,
while novel or difficult problems trigger high-entropy exploration and
explicit method acquisition. Verification experiments on the QSS60 benchmark
demonstrate that entropy-guided selection consistently yields superior
semantic coverage and greater internal diversity compared with a random
baseline, confirming the effectiveness of the proposed approach.

Future research will further explore how explicit method memory can be
integrated with LLMs in real-world applications such as IoT systems,
diagnostic tasks, and rare-event reasoning. In particular, evaluating the
robustness of high-entropy methods, automatically verifying their correctness,
and incorporating feedback-driven refinement represent promising directions.
Additional work may investigate hybrid architectures that combine parametric
and non-parametric learning more seamlessly, enabling lifelong learning,
dynamic method evolution, and greater interpretability in complex environments.

% \section*{Acknowledgment}
% The authors thanks th.

% Can use something like this to put references on a page
% by themselves when using endfloat and the captionsoff option.
\ifCLASSOPTIONcaptionsoff
  \newpage
\fi

\bibliographystyle{IEEEtran}
\bibliography{ref}

% biography section
%
% If you have an EPS/PDF photo (graphicx package needed) extra braces are
% needed around the contents of the optional argument to biography to prevent
% the LaTeX parser from getting confused when it sees the complicated
% \includegraphics command within an optional argument. (You could create
% your own custom macro containing the \includegraphics command to make things
% simpler here.)
%\begin{IEEEbiography}[{\includegraphics[width=1in,height=1.25in,clip,keepaspectratio]{mshell}}]{Michael Shell}
% or if you just want to reserve a space for a photo:

\begin{IEEEbiography}{Hong Su}
  received the MS and PhD degrees, in 2006 and 2022, respectively, from Sichuan University, Chengdu, China. He is currently a researcher of Chengdu University of Information Technology Chengdu, China. His research interests include blockchain, cross-chain and smart contract.
\end{IEEEbiography}

% You can push biographies down or up by placing
% a \vfill before or after them. The appropriate
% use of \vfill depends on what kind of text is
% on the last page and whether or not the columns
% are being equalized.

%\vfill

% Can be used to pull up biographies so that the bottom of the last one
% is flush with the other column.
%\enlargethispage{-5in}

% that's all folks
\end{document}